\documentclass[10pt,twocolumn,letterpaper]{article}

\usepackage{cvpr}              

\definecolor{cvprblue}{rgb}{0.21,0.49,0.74}
\usepackage[pagebackref,breaklinks,colorlinks,allcolors=cvprblue]{hyperref}

\def\paperID{11760} 
\def\confName{CVPR}
\def\confYear{2025}

\title{Towards More General Video-based Deepfake Detection through Facial 
Component Guided Adaptation for Foundation Model}

\usepackage{algorithm}
\usepackage{algpseudocode}
\newcommand{\textbt}[1]{\textbf{\textit{#1}}}
\usepackage{multirow}

\let\svthefootnote\thefootnote
\newcommand\blankfootnote[1]{%
  \let\thefootnote\relax\footnotetext{#1}%
  \let\thefootnote\svthefootnote%
}

\author{
{Yue-Hua Han$^{1,3,4}$,\ Tai-Ming Huang$^{1,3,4}$,\ Kai-Lung Hua$^{2,4}$,\ Jun-Cheng Chen$^{1}$}\\
{$^{1}$Academia Sinica $^{2}$Microsoft  $^{3}$National Taiwan University}\\
{$^{4}$National Taiwan University of Science and Technology}
}

\begin{document}
\maketitle


\begin{abstract}
Generative models have enabled the creation of highly realistic facial-synthetic images, raising significant concerns due to their potential for misuse. Despite rapid advancements in the field of deepfake detection, developing efficient approaches to leverage foundation models for improved generalizability to unseen forgery samples remains challenging. To address this challenge, we propose a novel side-network-based decoder that extracts spatial and temporal cues using the CLIP image encoder for generalized video-based Deepfake detection. Additionally, we introduce Facial Component Guidance (FCG) to enhance spatial learning generalizability by encouraging the model to focus on key facial regions. By leveraging the generic features of a vision-language foundation model, our approach demonstrates promising generalizability on challenging Deepfake datasets while also exhibiting superiority in training data efficiency, parameter efficiency, and model robustness.
\end{abstract}
\vspace{-10pt}

\section{Introduction}
\label{sec:intro}

With advancements in deep learning, generative models~\cite{ProGAN,Scale_up_GAN,stableDM,lora,StyleGAN2,DALLE2,GLIDE} have empowered the general public to create realistic synthetic images. In the wrong hands, these tools can easily be used to mislead viewers, incite social unrest and pose significant threats for personal reputation. In recent years, research on Deepfake detection has gained momentum to counter the growing risks posed by synthetic contents. However, many existing detection methods struggle to recognize Deepfakes, synthetic facial images and videos, generated by newer synthesis techniques~\cite{facexray,lipsdontlie,sbi,CDF,DFDC,3D-Deco}. This underscores the importance of developing models with robust generalization capabilities that can accurately differentiate between real and Deepfake samples under unfamiliar scenarios.

Previous works~\cite{facexray,lipsdontlie,sbi,sladd,caddm,realforensic,SLF,MLR,LSDA,AltFreeze,tall} have addressed the generalization problem from various angles, such as utilizing specific facial features for Deepfake detection~\cite{facexray,lipsdontlie}, generating pseudo-Deepfake samples for self-supervised learning~\cite{sbi,sladd}, detecting temporal inconsistencies in Deepfake videos~\cite{ftcn}, and fine-tuning pre-trained expert models~\cite{realforensic}. Despite these efforts, recent methods still face challenges, often learning dataset-specific cues that fail to detect unseen forgery samples.

\begin{figure}[t]
\centering
\includegraphics[trim={3.3cm 0cm 3.2cm 0cm},clip,width=0.95\linewidth]{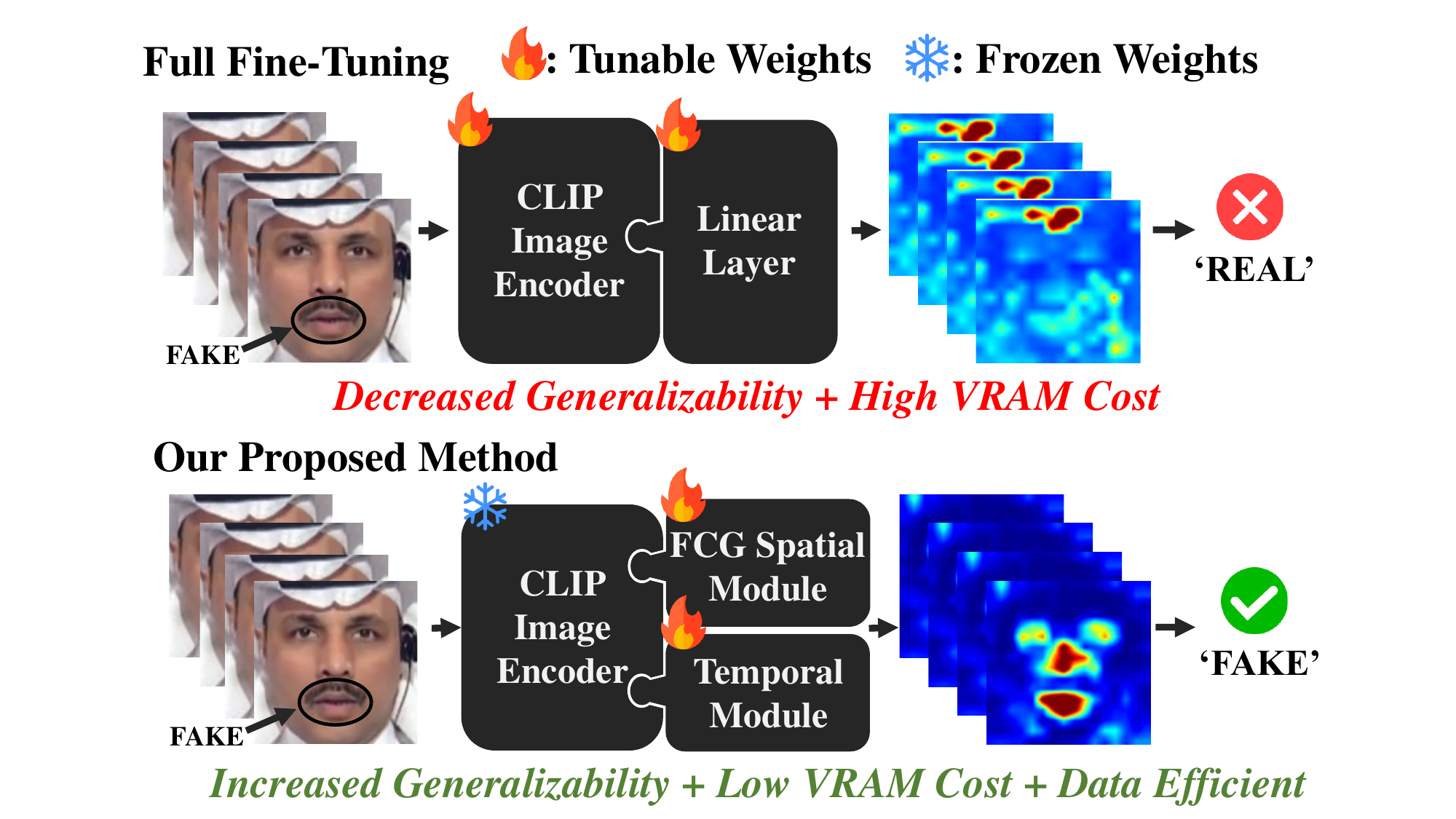} 
\caption{Our proposed method incorporates a Facial Component Guided (FCG) spatial module and a temporal module to adapt the foundational CLIP image encoder for the task of Deepfake detection. This approach enhances model generalizability by guiding it to focus on nuanced artifacts that appear in specific facial component regions rather than arbitrary areas.}
\label{fig:teaser}
\end{figure}

Foundation Models (FMs)~\cite{CLIP,dinov2,SAM} -- generic models trained on web-scale datasets -- have demonstrated remarkable zero-shot performance and can be quickly adapted to various downstream tasks with few additional tuning cost. In terms of general synthetic image detection, Ojha et al.~\cite{Ojha} has adapted the Contrastive Language-Image Pretraining (CLIP)~\cite{CLIP} with an additional linear layer and achieved an impressive performance. However, its effectiveness in detecting Deepfakes (synthetic facial images/videos) remains largely unexplored. From our pilot study as shown in Figure~\ref{fig:teaser}, the full fine-tuned CLIP-based detector tends to focus its attention on the generic regions and may potentially reduce its generalization performance. This raises an important question: How can we harness the rich inherent features of FMs to tackle the Deepfake detection task while benefiting from their advantages of fast adaptation and generalizability? 

To address the above challenges, inspired by recent advancements in Parameter Efficient Fine-tuning Techniques (PEFT), we propose an efficient and generalizable framework based on a side-network-based decoder -- comprising both spatial and temporal modules -- to extract generic spatio-temporal representation based on the layer-wise features from the CLIP image encoder for the Deepfake detection task. Furthermore, we develop a Facial Component Guidance (FCG) mechanism to enhance the generalizability of the spatial module by focusing on key facial components during training. With extensive experiments and ablation studies, the proposed method outperforms other State-Of-The-Art (SOTA) methods in the cross-dataset evaluation. The results demonstrate the effectiveness of our approach.

Our main contributions are summarized as follows:
\begin{itemize}
    \item We develop a novel side-network-based decoder with the spatial and temporal modules which not only extracts rich spatio-temporal features but also allows to fast adapt the CLIP image encoder to video-based Deepfake detection. 
    \item The proposed FCG mechanism effectively enhances the spatial module's generalization capability by directing the module's attention to key facial regions.
    \item Our method outperforms other SOTA methods in cross-dataset evaluation.
    Extensive experiments further demonstrate the efficiency of our framework in terms of the number of training parameters and data, as well as its robustness against generic perturbations.
\end{itemize}
 
\begin{figure*}[t]
\centering
\includegraphics[trim={2cm 1.2cm 0.5cm 0cm},clip,width=0.9\textwidth]{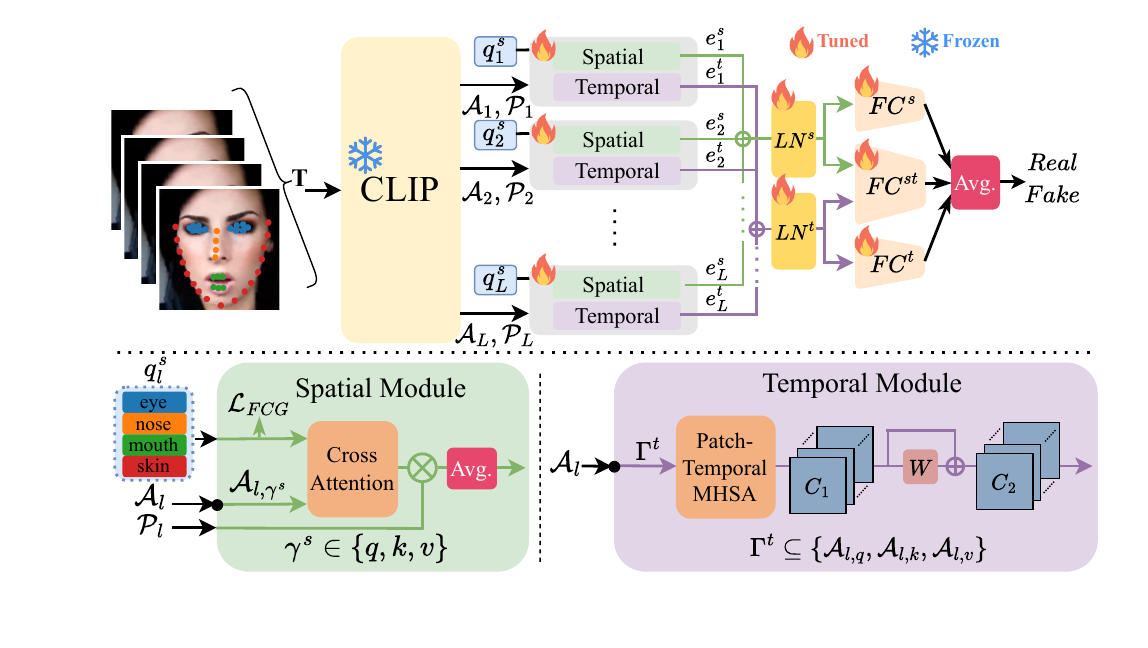} 
\caption{\textbf{Framework Overview:}  Our method utilizes the CLIP image encoder to extract layer-wise features (attention attributes $\mathcal{A}_l$ and patch embeddings $\mathcal{P}_l$), which are then processed by the corresponding decoder block consist of temporal and spatial modules for parameter-efficient fine-tuning. The spatial module incorporates the FCG loss to focus on key facial parts in each frame to capture the Deepfake visual cues. The temporal module employs the Patch-Temporal Multi-Head Self-Attention to capture the temporal inconsistency of Deepfake videos. The \(\bullet\) at the input of the two modules indicates where the required attributes are extracted based on the pre-configured settings.
Ultimately, our framework aggregates the outputs of the spatial and temporal modules for the final prediction. The superscripts $t$, $s$, and $st$ represent temporal related, spatial related, and spatio-temporal related components, respectively.}
\label{fig:structure}
\end{figure*}

\section{Related Work}
In this section, we briefly review recent works on (1) Foundation Models (FMs) and Parameter Efficient Fine-tuning Techniques (PEFTs), and (2) Deepfake detection.

\subsection{Foundation Model and Parameter Efficient Fine-tuning Techniques}
Adapting large-scale FMs for downstream tasks has become the new gold standard in the field of modern machine learning. These models are highly adaptable for various downstream tasks with superior performance, characterized by their training on large-scale datasets through advanced methods, which yields rich and powerful feature extraction capabilities. Leveraging multi-modal information further enhances these models’ semantic understanding. Notable examples include CLIP~\cite{CLIP} with its robust visual-text encodings, its expanded counterpart OpenCLIP~\cite{OpenCLIP}, the image-generating Stable Diffusion~\cite{stableDM}, and the self-supervised DINOv2~\cite{dinov2}. These models accelerate downstream task development, as demonstrated by EVL~\cite{frozen_clip} and DIFT~\cite{dift}, which utilize FMs for video action recognition and image correspondence, respectively.

Adapting these models to specific tasks poses significant challenges, particularly due to their large model sizes, which has led to the development of partial fine-tuning strategies. The Low-Rank Adaptation (LoRA)~\cite{lora} method adapts large language models by incorporating rank decomposition matrices into the Transformer structure, significantly reducing the number of trainable parameters and enabling efficient adaptation.\ The Visual-Prompt Tuning (VPT)~\cite{vpt} technique fine-tunes Vision Transformer (ViT) models using trainable ``prompts'' to minimize training overhead.\ The EXPRES~\cite{EXPRES} approach constructs downstream representations with residual learnable ``output'' tokens.\ The SAN~\cite{side_net} framework incorporates a streamlined side network into the static CLIP model for open-vocabulary semantic segmentation. Selecting the most effective adaptation strategy to maximize a model's capabilities remains an important research focus.

\subsection{Deepfake Detection}
The research and development of Deepfake detection have led to the exploration of a wide range of methods to address this challenge. In the early stages, Xception~\cite{ff++} showed strong performance in intra-dataset evaluations but suffered a significant performance drop in cross-dataset evaluations. Subsequent studies have focused on exploiting imperfections that generative models struggle to eliminate. For instance, FTCN~\cite{ftcn} targets temporal anomalies, LipsForensics~\cite{lipsdontlie} addresses unnatural lip motions, and EyesTellAll~\cite{eyestellall} detects irregularities in pupil shapes.

The issue of insufficient forgery sample diversity in the training dataset leading to poor model generalizability also remains an important problem. Recent studies~\cite{sbi,facexray,sladd,pcli2g,LAA-Net} have incorporated self-supervised learning to synthesize pseudo-Deepfake samples to tackle such problem. RealForensics~\cite{realforensic} also pre-trains its video backbone on large lip-reading datasets and further transfer the model to Deepfake detection. In ~\cite{LSDA}, Yan et al. propose to simulate forgery features in the latent space to support detectors to learn a more generalizable decision boundary. 

Further studies on video-based Deepfake detection have emerged to address challenges in generalizability and memory efficiency. AltFreezing~\cite{AltFreeze} alternately freezes spatial and temporal weights of a network during training. TALL-Swin~\cite{tall} transforms video clips into predefined layouts to preserve spatial and temporal dependencies for Deepfake detection. In~\cite{SLF}, the authors leverage the style latent flow from StyleGAN~\cite{StyleGAN3} to enhance the generalizability of Deepfake video detectors. Other notable efforts to improve model generalizability include addressing the identity leakage problem~\cite{caddm} and simultaneously tackling both binary Deepfake classification and manipulation order detection~\cite{MLR}. However, the effectiveness of FMs in this field remains largely unexplored.

\section{Method}
Our approach focuses on effectively harnessing the inherent capabilities of foundation models to identify temporal inconsistencies and spatial manipulations in Deepfake videos. To this end, we develop a framework that leverages the rich representations of the pre-trained foundation model (i.e., CLIP image encoder) and integrates a side-network-based decoding block, comprising both specialized temporal and spatial modules, to capture these key characteristics for Deepfake detection, as illustrated in Figure~\ref{fig:structure}. In the following sections, we first outline the model's overall architecture (Section~\ref{sec:overall}), and then delve into the details of the proposed {spatial (Section~\ref{sec:spatial}) and temporal (Section~\ref{sec:temporal}) modules in the layer-wise decoders}.

\begin{figure*}[th]
\centering
\includegraphics[width=0.95\textwidth]{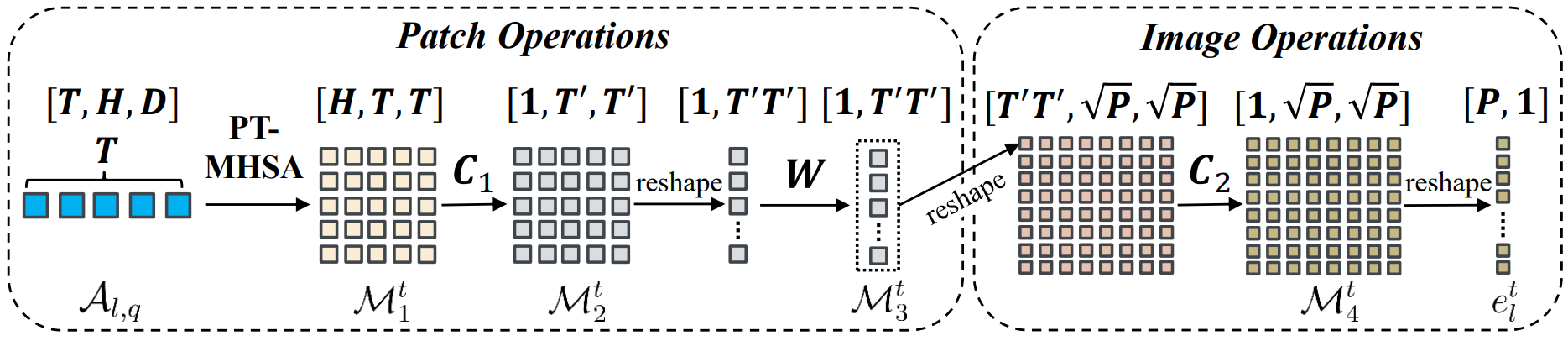} 
\caption{\textbf{Illustration of the Temporal Module Mechanism:} For enhanced clarity, we present a step-by-step demonstration of the operation within the temporal module. For illustration purpose, this example only considers a single attribute, where \(\Gamma^{t}= \{\mathcal{A}_{l,q}\}\). For scenarios involving multiple attributes, simply replace \(H\) with (\(|\Gamma^{t}| \times H\)).}
\label{fig:t_module}
\end{figure*}

\subsection{Overall Framework}\label{sec:overall}
Drawing inspiration from the works of \cite{frozen_clip,STAN}, we utilize the ViT-based CLIP image encoder to extract visual features that our side-network-based spatial and temporal modules can leverage for Deepfake detection. To facilitate understanding of the proposed method, we first introduce several symbols and notations representing the system configuration: \(L\) denotes the number of transformer layers, \(P\) the number of patches, \(T\) the number of input frames, \(H\) the number of attention heads in multi-head self-attention, and \(D\) the feature dimension per head. Besides, \textbf{we follow the convention throughout the paper that the superscripts $t$, $s$, and $st$ denote temporal, spatial, and spatio-temporal components, respectively. (e.g., \(x^{t}\), \(x^{s}\), and \(x^{st}\).)}

For each transformer layer of the image encoder, we obtain patch embeddings \( \mathcal{P}_l \in \mathbb{R}^{T \times P \times (H \times D)} \) and attention attributes \( \mathcal{A}_{l,\gamma} \in \mathbb{R}^{T \times P \times H \times D} \), where \( 1 \le l \le L \) indicates the layer index, and \( \gamma \in \{q,k,v\} \) specifies the attention attribute type (i.e., query, key and value). Please refer to Section~\ref{sec:elab_attn_attr_extract} for more details. These layer-wise attributes are then processed by a layer-wise decoder block through the spatial and temporal modules, resulting in temporal and spatial embeddings \( e^{t}_l \) and \(e^{s}_{l}\). We aggregate these layer-wise temporal and spatial embeddings, denoted as \(E^{t}\) and \(E^{s}\), which are subsequently normalized using respective LayerNorm layers \(\textbt{LN}^{t}\) and \(\textbt{LN}^{s}\). The framework employs three independent Fully-Connected (FC) classification heads to predict the level of Deepfake in the  temporal \(\textbt{FC}^{t}\), spatial \(\textbt{FC}^{s}\), and spatio-temporal \(\textbt{FC}^{st}\) branches. The overall Deepfake score is the average score of the respective heads.

\subsection{Spatial Module}\label{sec:spatial}
The spatial module in our decoder block is designed to capture manipulation cues from the extracted patch features \( \mathcal{P}_l \). In the following, we first introduce the spatial module pipeline, then we describe the mining process used to extract layer-wise facial component attributes for the FCG. Finally, we detail the FCG loss used during training, which steers the spatial module's focus towards key facial regions such as the lips, skin, eyes, and nose.

\vspace{5pt}
\noindent \textbf{Pipeline.} Each spatial module in the decoder block has its own set of \(N\) learnable parameters, denoted as \( \textbt{q}^{s}_l \in \mathbb{R}^{N \times (H \times D)}\). These learnable parameters are fed into the Cross-Attention (CA) module as queries, while the keys are the target attribute \(\gamma^{s}\) of the video patches, and the values are the video patch embeddings. This process is expressed as \( \mathcal{M}^{s}_{1} = \textbt{CA}( \textbt{q}^{s}_l, \mathcal{A}_{l,\gamma^{s}},  \mathcal{P}_l ) \), where \( \mathcal{M}^{s}_1 \in \mathbb{R}^{T \times N \times (H \times D)} \), 
followed by averaging over the temporal and query dimensions of \( \mathcal{M}^{s}_1 \) and reshaping the tensor to obtain 
the layer-wise spatial feature \( e^{s}_{l} \in \mathbb{R}^{(H \times D)\times 1}\). 

\vspace{5pt}
\noindent \textbf{Facial Attribute Mining.} In this step, we collect layer-wise features associated with key facial parts (lips, skin, eyes, nose) from the CLIP image encoder to further guide the learnable queries \( \textbt{q}^{s}_l \) in the spatial module during training. First, we gather frames from facially aligned videos with random augmentations, such as resizing, cropping, and bidirectional flipping. These frames are then fed into the CLIP image encoder to extract the target layer-wise attributes \( \mathcal{A}_{l,\gamma^{s}} \). Facial landmarks are detected for each frame, allowing us to collect the patch attributes corresponding to the facial part landmarks. The patch attributes for each facial part are averaged after \( L_2 \)-normalization, producing the layer-wise facial part attributes \( \varphi^{s}_{l} \in \mathbb{R}^{4 \times (H \times D)} \), where the first dimension corresponds to the lips, skin, eyes, and nose, respectively. The augmentations applied in the first step are crucial to remove the positional encoding in the features. 

\vspace{5pt}
\noindent \textbf{Facial Component Guidance.} 
When applying the Facial Component Guidance (FCG), the number of parameters \(N\) in the spatial module is constrained to align with the number of facial parts {($1 \le N \le 4$)}. The FCG loss is defined as the InfoNCE loss between \( \textbt{q}^{s}_l \) and \( \varphi^{s}_{l} \), encouraging each query in \( \textbt{q}^{s}_l \) to closely match its corresponding facial part attribute in \( \varphi^{s}_{l} \). The FCG loss is computed as shown in Equation \ref{eq:loss_guide}:
\begin{equation}
\label{eq:loss_guide}
{\mathcal{L}}_{FCG} =\frac{1}{NL}\sum^{L}_{l=1}{\sum^{N}_{i=1}{\ln\left(\frac{\exp(\textbt{cos}(q^{s}_{l,i},\varphi^{s}_{l,i})\times\tau)}{\sum^{N}_{j=1}{\exp(\textbt{cos}(q^{s}_{l,i},\varphi^{s}_{l,j})\times\tau)}}\right)}},
\end{equation}
\noindent where \( \tau \) denotes the softmax temperature and \( \textbt{cos}(\cdot, \cdot) \) stands for the cosine similarity between two vectors. We apply the InfoNCE loss to contrastively guide the learnable queries, aiming to balance freedom between learning forgery cues and maintaining generalizability.

\subsection{Temporal Module}\label{sec:temporal}

Our temporal module in the decoder blocks focuses on capturing the temporal inconsistencies in Deepfake samples. 
As illustrated in Figure~\ref{fig:t_module}, it consists of four major steps, which are described below in detail.

For each fixed patch location, we apply Multi-Head Self-Attention (MHSA) along the temporal axis of the attributes in \(\Gamma^{t} \subseteq \{\mathcal{A}_{l,q},\mathcal{A}_{l,k}, \mathcal{A}_{l,v}\} \), we refer this process as Patch-Temporal Multi-Head Self-Attention (PT-MHSA). This step creates a temporal affinity map \( \mathcal{M}^{t}_{1} \in \mathbb{R}^{P \times (|\Gamma^{t}| \times H) \times T \times T} \). We then apply the softmax function along the last axis to normalize the affinity values.

Next, we aggregate the multi-attribute, multi-head affinity scores at adjacent temporal locations using a 2D kernel \( \textbf{C}_{1} \). This kernel convolves over the last two axes of \( \mathcal{M}^{t}_{1} \), with the second axis (\(|\Gamma^{t}| \times H\)) as the input channel, producing a feature map \( \mathcal{M}^{t}_{2} \in \mathbb{R}^{P \times T' \times T'} \).

To promote interaction across all temporal representations, we reshape the last two axes of \(\mathcal{M}^{t}_{2}\) into a single axis with dimensions \(T' \times T'\), then process the feature map through a residual linear layer \(\textbf{W}\) to produce a new feature map \(\mathcal{M}^{t}_{3} \in \mathbb{R}^{P \times (T' \times T')} \).

Finally, to aggregate with spatially adjacent patches, we reshape \( \mathcal{M}^{t}_{3} \) into \( \mathbb{R}^{(T' \times T') \times \sqrt{P} \times \sqrt{P}} \) and apply a 2D convolution kernel \( \textbf{C}_{2} \) to create the feature map \(\mathcal{M}^{t}_4 \in \mathbb{R}^{\sqrt{P} \times \sqrt{P}}\). This feature map is reshaped into \(e^{t}_{l} \in \mathbb{R}^{P \times 1}\), serving as the temporal feature for the \(l\)-th decoder layer.

\subsection{Objective Functions}
The main objective comprises the focal loss~\cite{focal} \( \mathcal{L}_{F}(\cdot) \) of the three classification heads for temporal (${\mathcal{L}}_{T}$), spatial (${\mathcal{L}}_{S}$), and spatio-temporal (${\mathcal{L}}_{ST}$) aspects:
\begin{align}
\label{eq:loss_T}
{\mathcal{L}}_{T} &= \mathcal{L}_{F}(\textbt{FC}^{t}(\textbt{LN}^{t}(E^{t})),y), \\
\label{eq:loss_S}
{\mathcal{L}}_{S} &= \mathcal{L}_{F}(\textbt{FC}^{s}(\textbt{LN}^{s}(E^{s})),y), \\
\label{eq:loss_ST}
{\mathcal{L}}_{ST} &= \mathcal{L}_{F}(\textbt{FC}^{st}(\textbt{cat}(\textbt{LN}^{t}(E^{t}),\textbt{LN}^{s}(E^{s}))),y),
\end{align}
\noindent where \( \textbf{cat}(\cdot) \) represents the concatenation operation that merges the given features, \(E^{s}\) and \(E^{t}\) are spatial and temporal features aggregated from each layer, and \(y\) is the ground truth binary label. The overall objective during training is defined as:
\begin{equation}
\label{eq:loss_all}
 \mathcal{L}_{total} = \mathcal{L}_{T} + \mathcal{L}_{S} + \mathcal{L}_{ST} + w\mathcal{L}_{FCG},
\end{equation}
\noindent with \(w\) adjusting the influence of the FCG loss.

\section{Experiments}
In the following sections, we present the implementation details, introduce the evaluation datasets and settings, and provide extensive experiments to demonstrate the effectiveness of our framework across various aspects, including generalizability, parameter efficiency, data efficiency, and robustness. For additional experiments please refer to the supplementary materials.

\begin{table}[t]
\small
\setlength{\tabcolsep}{1pt} 
\renewcommand{\arraystretch}{1.2} 
\begin{center}
\begin{tabular}{  l | c | c c c c c }
\hline
Model   & Pub. / Mod. & CDF & DFDC &FSh & DFo & WDF \\
\hline
Xception~\cite{ff++}  & {\footnotesize ICCV'19 / Img.} & 73.7  & 70.9  & 72.0 &  84.5 & -  \\
Face X-ray~\cite{facexray} &  {\footnotesize CVPR'20 / Img.}  & 79.5  & 65.5   & 92.8 & 86.8  & - \\
LipForensics~\cite{lipsdontlie} & {\footnotesize CVPR'21 / Vid.}  & 82.4  & 73.5   & 97.1 & 97.6 & -\\
FTCN~\cite{ftcn} & {\footnotesize ICCV'21 / Vid.}  & 86.9 & 74.0 & 98.8 & 98.8 & - \\
SBI(c23)~\cite{sbi} & {\footnotesize CVPR'22 / Img.}  & 92.9 & 72.0 & - & - & -  \\
RealForensics~\cite{realforensic} & {\footnotesize CVPR'22 / Vid.}  & 86.9  & 75.9 & \textbf{99.7} & \underline{99.3}   &  - \\
AltFreezing~\cite{AltFreeze} & {\footnotesize CVPR'23 / Vid.}   & 89.5 & - & \underline{99.4} & \underline{99.3} & - \\
CADDM~\cite{caddm} & {\footnotesize CVPR'23 / Img.}  & \underline{93.9}  & 73.9 & - &  -   & - \\
TALL-Swin ~\cite{tall} & {\footnotesize ICCV'23 / Vid.}   & 90.8  & \underline{76.8} & \textbf{99.7} & \textbf{99.6}   & - \\
Yan et al.~\cite{LSDA} & {\footnotesize CVPR'24 / Img.}   & 91.1  & \textbf{77.0} & - &  -   & -\\
Choi et al.~\cite{SLF} & {\footnotesize CVPR'24 / Vid.}   & 89.0 & - & 99.0 &  99.0   & - \\
Hong et al.~\cite{MLR} & {\footnotesize CVPR'24 / Img.}   & 91.6  & 75.2 & - &  -   & \underline{73.4} \\
LAA-Net~\cite{LAA-Net} & {\footnotesize CVPR'24 / Img.}  & \textbf{95.4} & - & - & - & \textbf{80.0} \\
\hline
\hline
RealForensics* & {\footnotesize CVPR'22 / Vid.}  & 88.9  & \underline{75.8} & \textbf{98.7} & \underline{99.5}  &  \underline{83.3} \\
LAA-Net* & {\footnotesize CVPR'24 / Img.}   & \underline{93.6} & 73.8 & 69.8 & 75.5 & 80.8 \\

\hline 
\textbf{Ours} & -  &  \textbf{95.0}  & \textbf{81.8} & \underline{98.1} & \textbf{99.6} &   \textbf{87.2} \\
\hline
\end{tabular}
\caption{\textbf{Generalizability Evaluation on Unseen Datasets}: The video-level AUC (\%) of our method on unseen datasets (CDF, DFDC, FSh, DFo, and WDF) after training on the FF++ dataset. The scores from previous works are directly reported from their respective papers. Models marked with * are re-evaluated using the official code and model checkpoint under default settings. {The publication and modality of each work are provided for reference.}}
\label{table:cross_dataset_experiment}
\end{center}
\vspace{-10pt}
\end{table}

\subsection{Implementation Details}\label{sec:implementation}

We adopt a similar video pre-processing pipeline to RealForensics~\cite{realforensic}, which includes the following steps: 1) utilizing 2D-FAN\footnote{https://github.com/1adrianb/face-alignment} to extract facial landmarks from video frames, 2) aligning the detected faces with the mean face from the LRW~\cite{LRW} dataset, and 3) extracting the aligned faces from the videos.  
During training, non-overlapping clips ranging from 2 to 4 seconds are randomly extracted from each video, and 10 frames are uniformly sampled (\(T=10\)) from the extracted clips. The sampled frames undergo a video-level augmentation pipeline, which includes horizontal flipping, random resizing and cropping, RGB shifting, hue and saturation adjustments, random modifications to brightness and contrast, image compression, blurring, and downscaling. This diverse augmentation strategy aims to simulate a wide range of real-world variations.  

Unless otherwise specified, our framework is based on the CLIP ViT-L/14\footnote{https://github.com/openai/CLIP} image encoder, which consists of 24 encoder layers (\(L=24\)). In the spatial module, we set \( {\gamma^{s}} = k\). For the temporal module, we set \(\Gamma^{t} = \{\mathcal{A}_{l,q},\mathcal{A}_{l,k}, \mathcal{A}_{l,v}\} \) and the convolution kernels \(\mathcal{C}_1\) and \(\mathcal{C}_2\) are configured with a kernel size of 5, padding of 2, and a stride of 1. We employ the AdamW optimizer, setting the learning rate (\(lr\)) to 1e-4, with \(\beta\) parameters set to (\(0.9, 0.999\)) and a weight decay of 1e-3. The weight (\(w\)) of the loss function is set to 0.15. We utilize focal loss with a gamma value of 4 to focus on the challenging samples. To optimize training, we implement an early stopping strategy, monitoring the Area Under the Receiver Operating Characteristic (AUROC) score on the validation dataset with a patience of 10 epochs. Our training process spans 30 epochs using \(4 \times \text{V100}\) GPUs, with a batch size of 60 clips per GPU. \textbf{The overall training process is completed within approximately 16 GPU hours}.

\subsection{Evaluation Settings for Deepfake Detection}
Aligning with all prior work in Deepfake detection is challenging due to the diversity of evaluation protocols. For fair comparison, we primarily follow the evaluation protocols established by RealForensics~\cite{realforensic}. The datasets used in our experiments are as follows: The \textbf{FaceForensics++ (FF++)} dataset~\cite{ff++} contains 1,000 real videos and 4,000 Deepfake videos generated with face-swapping techniques (Deepfake~\cite{Deepfake}, FaceSwap~\cite{FaceSwap}) and reenactment methods (NeuralTexture~\cite{NT}, Face2Face~\cite{F2F}). The \textbf{CelebDF-v2 (CDF)}~\cite{CDF} dataset includes 5,639 high-quality DeepFake videos of celebrities. The \textbf{DeepFake Detection Challenge (DFDC)}~\cite{DFDC} dataset is a large-scale dataset containing 124,000 Deepfake videos generated with eight different algorithms. The \textbf{FaceShifter (FSh)}~\cite{FSh} dataset produces high-fidelity Deepfake videos using an occlusion-aware face swapping algorithm. The \textbf{DeeperForensics (DFo)}~\cite{DFo} dataset comprises 60,000 videos with generic perturbations to enhance data diversity. Lastly, the \textbf{WildDeepfake (WDF)}~\cite{WDF} dataset includes 7,314 face sequences from 707 Deepfake videos sourced from the internet. We use the complete FF++ dataset for our experiments, but we only use the test split of the other datasets for evaluation purpose.\footnote{Following RealForensics, we use only a subset (3,814 samples) of the DFDC test split (5,000 samples) for evaluation.}

Unless otherwise stated, the experiments follow the given regime: 1) The models are trained on the training subset of the High Quality (HQ) compression FF++ dataset. 2) The reported scores are presented in video-level AUROC in percentage. 3) Values in \textbf{bold} represent the best results, while those \underline{underlined} indicate the second-best results.

\begin{table}[t]
\setlength{\tabcolsep}{3pt} 
\renewcommand{\arraystretch}{1} 
\begin{center}
\begin{tabular}{  l | c c c c c| l  }
\hline
Comp.  &    CDF & DFDC &FSh & DFo & WDF & Avg.\\
\hline
\textbf{Ours}  & \textbf{95.0}  & \underline{81.8} & \underline{98.1} & \textbf{99.6} & \textbf{87.2} &   \textbf{92.3} \\
\hline
w/o S mod.	& 86.2 & 75.7 &	\textbf{98.3} & \underline{99.4} & 75.2 & 87.0(-5.3) \\
w/o T mod.	& 91.0 & \textbf{83.0} &	87.0 & 98.0 & \underline{86.1} & 89.0(-3.3) \\
w/o FCG		& \underline{92.0} & 81.7 & 96.3 & 99.1 & 85.2 & 90.9(-1.4) \\
w/o focal	& 91.9 & 81.5 & \underline{98.1} & 99.0 & 85.5 & \underline{91.2}(-1.1) \\
\hline 
\end{tabular}
\caption{\textbf{Ablation Study}: The components are the spatial module (S mod.) and the temporal module (T mod.) within the decoder block, the proposed FCG loss (FCG) and the focal loss (focal).}
\label{table:ablate_guide_components}
\end{center}
\end{table}

\begin{table}[t]
\begin{center}
\setlength{\tabcolsep}{2.5pt} 
\renewcommand{\arraystretch}{1} 
\begin{tabular}{  l | c || c || c c c c c| c}
\hline
 w/o & Model  & FF++ & CDF & DFDC & FSh & DFo & WDF & Avg. \\
\hline
\multirow{2}{1em}{DF} & R.F.   & \textbf{100.0} & 69.2 & 72.4 & 90.5 & \textbf{99.2}& 54.2 &  77.1\\
& Ours       & 99.6 & \textbf{93.0} & \textbf{79.1} & \textbf{97.1} & \textbf{99.2}& \textbf{83.1} & \textbf{90.3} \\
\hline
\multirow{2}{1em}{FS} & R.F.   & \textbf{97.1} & 80.0 & 71.2 & \textbf{98.5} & 99.4 & 78.8 & 85.6\\
& Ours       & 93.2 & \textbf{93.2} & \textbf{75.9} & 96.3 & \textbf{99.5} & \textbf{86.8} & \textbf{90.3}\\

\hline
\multirow{2}{1em}{F2F} & R.F.   & 99.7 & 75.7 & 72.3 & \textbf{99.1} & \textbf{99.9} & \textbf{83.1} & 86.0\\
& Ours & \textbf{99.8} & \textbf{92.5} & \textbf{81.1} & 96.6 & 99.0  & 82.3& \textbf{90.3} \\

\hline
\multirow{2}{1em}{NT} & R.F.   & \textbf{99.2} & 80.8 & 72.4 & 93.8 & 98.6 & 68.3 & 82.8\\
& Ours       & 94.6 & \textbf{92.4} & \textbf{82.3} & \textbf{96.4} & \textbf{99.2} & \textbf{84.4} & \textbf{90.9}\\
\hline

\end{tabular}

\caption{\textbf{Extended FF++ Leave-One-Out (LOO) Evaluation:} In this experiment, we compare with RealForensics (R.F.) by extending the FF++ LOO evaluation to include cross-dataset assessment. The R.F. performance on the FF++ dataset is directly reported from the paper, while the scores for cross-datasets are evaluated using the LOO weights provided by the official repository. }
\label{table:leave_one_out_experiment}
\end{center}
\vspace{-10pt}
\end{table}

\begin{table}[t]
\setlength{\tabcolsep}{2.5pt} 

\begin{center}

\begin{tabular}{l| r | c c c c c| l }
\hline
Method & \#  T.P. & CDF & DFDC & FSh & DFo & WDF & Avg. \\
\hline
LCP  & 2.0K & 75.8& 77.0 & 88.0 & 92.4 & 82.1 & 83.3 \\ 
\(\text{VPT}_\text{Shallow}\)  & 3.1K & 78.9 & 76.4 & 89.0 & 92.7 & 82.9 & 84.0  \\
\(\text{VPT}_\text{Deep}\)  & 26.6K & 81.1& \underline{77.5} & \underline{90.0} & \underline{98.3} & \underline{86.7} & \underline{86.7}  \\
EVL   & 58.3M & 80.3 & 75.8 & 87.7 & 96.5 & 79.1 & 83.9 \\ 
FFT  & 303.0M & \underline{88.3}& 75.2 & 84.8 & 93.8 & 83.8 & 85.1  \\
\hline
Ours &  250.0K & \textbf{95.0}  & \textbf{81.8} & \textbf{98.1} & \textbf{99.6} &  \textbf{87.2} &\textbf{92.3} \\
\hline
\end{tabular}
\caption{\textbf{Comparison of Model Architectures:} We compare our adaptation methods with the various adaptation methods under the cross-dataset evaluation. The models are based on the ViT-L/14 CLIP image encoder. We also present the number of Trainable Parameters (\# T.P.) for comparison. ~\ref{sec:implementation}.
}
\label{table:cross_arch}
\end{center}
\end{table}

\begin{table}[t]
\begin{center}
\setlength{\tabcolsep}{6pt} 
\renewcommand{\arraystretch}{1} 

\begin{tabular}{  r |  c c c c c | c}
\hline
 \#Data  & CDF & DFDC & FSh & DFo & WDF & Avg. \\
\hline
100\% & \textbf{95.0} & \textbf{81.8} & \textbf{98.1} & \textbf{99.6} & \textbf{87.2} & \textbf{92.3} \\
75\% & \underline{93.1} & \underline{81.4} & \underline{97.7} & 99.3 & \underline{85.5} & \underline{91.4} \\
50\% & 92.0 & \underline{81.4} & 96.2 & \underline{99.4} & 84.7 & 90.7 \\
25\% & 90.3 & 77.2 & 95.6 & 99.1 & 80.1 & 88.5 \\
10\% & 78.3 & 73.3 & 91.9 & 99.0 & 75.2  & 83.5 \\
\hline
\end{tabular}
\caption{\textbf{Generalizability on Proportional Training Dataset:} To validate the efficiency of our approach on training samples, we assess the generalizability of our model trained on varying proportions of the training dataset. This evaluation demonstrates the model's performance in scenarios with limited training samples.}
\label{table:partial_dataset}
\end{center}
\vspace{-10pt}
\end{table}

\begin{table*}[th!]
\setlength{\tabcolsep}{4pt} 
\renewcommand{\arraystretch}{1} 
\begin{center}
\begin{tabular}{l|c |c|ccccccc|l}
\hline
Method  & Dataset & Clean &  CS & CC & Block & Noise & Blur & JPEG & Comp. & Avg \\
\hline
RealForensics & \multirow{2}{4em}{FF++~\cite{ff++}} & 99.8  & 99.8 & 99.6 & 98.9 & 79.7 & 95.3 & 98.4 & 97.6 & 95.6 \textbf{(-4.2)} \\
Ours\(^\dagger\) & & 99.2 & 97.9 & 97.5 & 98.1  & 77.1  &  94.3 &  94.1 & 93.4 & 93.2 (-6.0) \\
\hline
RealForensics\(^*\) & \multirow{2}{4em}{CDF~\cite{CDF}}  & 79.3  & 71.5   & 64.8  & 68.3  & 58.8  & 56.6  & 62.9  & 63.4  & 63.8 (-15.5) \\
Ours\(^\dagger\) &  & 91.0 & 89.4 & 87.6  &  90.1 & 74.6  & 77.9  & 80.1  &  75.7 & 82.2 \textbf{(-8.8)} \\
\hline
\end{tabular}
\caption{\textbf{Video Robustness Evaluation:} The upper part of the table reports the robustness evaluation on FF++, while the lower part reports the evaluation on CDF. We follow the training protocols mentioned in Section \ref{sec:robustness} and denote the model as Ours\(^\dagger\). We report the RealForensics' FF++ robustness evaluation directly from the paper, while the robustness evaluation on CDF is conducted with the official checkpoint (denoted as RealForensics\(^*\)). }
\label{table:robustness}
\end{center}
\vspace{-10pt}
\end{table*}

\subsection{Generalization to Unseen Dataset}\label{sec:cross_dataset}
In Table \ref{table:cross_dataset_experiment}, we assess the generalization capability of our model trained on the FF++ dataset against five unseen datasets: CDF, DFDC, FSh, DFo and WDF. We include scores from previous methods as reported in their respective papers and official repositories. By integrating CLIP's robust generalizability with our proposed framework, we observe notable improvements while facing unseen samples, outperforming the SOTAs on most datasets. This improvement is particularly pronounced in the challenging CDF, DFDC and WDF datasets, where our method establishes a significant lead of 1.4\%, 6\% and 3.9\%, respectively.

\subsection{Ablation Study on Model Components}

In Table \ref{table:ablate_guide_components}, we evaluate the contribution of various components to the framework's generalization capability across different datasets. Our experiment highlights the crucial role of our dual-module architecture: removing either the spatial or temporal module results in an average performance decline of 5.3\% and 3.3\%, respectively. Notably, excluding the spatial module significantly impacts performance on the challenging CDF, DFDC, and WDF datasets, underscoring the importance of facial semantic learning in accurately assessing video authenticity for these datasets. Additionally, removing the temporal module leads to a marked performance decrease on the FSh dataset, indicating the presence of temporal inconsistencies that provide important cues in this dataset. Our strategy of deploying independent modules to address both spatial and temporal aspects has proven effective for enhancing generalization.

Furthermore, we observe that using focal loss can slightly improve generalizability compared to cross-entropy loss, likely due to the focal loss's ability to handle hard samples. Our proposed Facial Component Guidance (FCG) further enhances generalization by directing the layer-wise queries to focus on primary facial regions, yielding an average performance boost of 1.4\%. The integration of these components collectively results in a significant improvement in the model's overall performance.

\subsection{Data-Efficiency Evaluation}
We further assess the generalizability of our framework under additional constraints on the training dataset, such as excluding certain types of manipulations and limiting the size of the dataset. These constraints evaluate the data efficiency of our framework in transferring knowledge in the CLIP image encoder to the Deepfake detection task. \\

\noindent \textbf{Constraint on FF++ Manipulation Types.}
\label{sec:loo_const}
In Table \ref{table:leave_one_out_experiment}, we follow the Leave-One-Out (LOO) protocol to evaluate the framework's performance on each of the four manipulation types in FF++ after training on the remaining three types. The results are presented under the FF++ column. Additionally, we assess the generalizability of the LOO model to unseen datasets (CDF, DFDC, FSh, DFo and WDF).

When evaluated on the FF++ dataset, RealForensics generally outperforms our proposed framework. However, RealForensics fails to maintain its performance in cross-dataset evaluations. In contrast, our approach consistently excels on the challenging CDF, DFDC, and WDF datasets and performs comparably on the FSh and DFo datasets. Since the DFo and FSh datasets use real videos from FF++ to create their Deepfake samples, we speculate that RealForensics may be biased towards FF++-specific cues for Deepfake detection. This could explain its stable performance on the FSh and DFo datasets, despite the significant performance drop when tested on the completely unseen CDF, DFDC, and WDF datasets. \\

\noindent\textbf{Constraint on Size of the Training Dataset.}
In this experiment, we scale the size of the FF++ training dataset to evaluate the framework's efficiency with a limited number of Deepfake samples. The results are presented in Table \ref{table:partial_dataset}, where the framework is trained on a portion of the dataset and then evaluated on the unseen datasets. Notably, our method exhibits robust performance, showing no significant decrease when trained on 75\% of the dataset. Furthermore, it surpasses state-of-the-art methods on most datasets even when trained with only 50\% of the dataset.

\subsection{Evaluation on Parameter Efficiency}
In Table~\ref{table:cross_arch}, we highlight the effectiveness and parameter efficiency of our proposed framework on adapting the CLIP image encoder for Deepfake detection, we compare our framework with various adaptation methods: the Full Fine-Tuning (FFT) method, the Linear Classifier Probing (LCP)~\cite{lprobe} method, the Visual Prompt Tuning (VPT)~\cite{vpt} method and the Efficient Video Learner (EVL)~\cite{frozen_clip} framework. We apply both the `Deep' and `Shallow' versions of VPT, where \(\text{VPT}_{\text{Deep}}\) introduces one trainable prompt per layer in the image encoder, while \(\text{VPT}_{\text{Shallow}}\) introduces a single trainable prompt only at the first layer.
Our framework is similar to EVL in deploying a side-network video learner that progressively decodes per-layer features. However, our framework employs two distinctive components with strong inductive bias -- the spatial and temporal modules -- unlike EVL, which uses an unified module to optimize all in a box. The experiment reveals our approach significantly outperforms previous adaptation methods across all datasets. Furthermore, our framework requires significantly fewer trainable parameters compared to EVL, being \(\approx\) 200 times smaller. This highlights our method's parameter efficiency and effectiveness in adapting the CLIP image encoder for Deepfake detection.

\subsection{Robustness Evaluation Against Perturbations}
\label{sec:robustness}
To assess the robustness of our framework against common perturbations, we follow the protocol described in RealForensics~\cite{realforensic}, training on the FF++ dataset with random flipping and random resize cropping augmentations. We then evaluate the model's performance on the FF++ testing subset with perturbations introduced by DeeperForensics (DFo)~\cite{DFo}. These perturbations vary across five intensity levels and affect the image in terms of color saturation (CS), color contrast (CC), block-wise occlusions (Block), color component noise (Noise), Gaussian blur (Blur), JPEG compression (JPEG), and video compression (Comp.). The results are summarized in the upper section of Table \ref{table:robustness}.

Our framework appears less robust than RealForensics on the FF++ dataset, showing an average performance drop of 1.8\% more when handling samples with perturbations. However, we note that RealForensics demonstrates strong affinity for the FF++ dataset, as discussed in Section \ref{sec:loo_const}. To provide a more comprehensive evaluation, we extend the robustness assessment to the unseen CelebDF (CDF)~\cite{CDF} dataset. We evaluate our model and the checkpoint provided by RealForensics\footnote{\href{https://github.com/ahaliassos/RealForensics?tab=readme-ov-file\#robustness}{Model checkpoint from RealForensics.}} on the CDF dataset with DFo-introduced perturbations applied. The results are presented in the lower section of Table \ref{table:robustness}. In this experiment, RealForensics struggles to maintain robustness when transitioning from FF++ to CDF, with an average performance drop of 15.5\%. In contrast, our framework exhibits a relatively stable performance with an average drop of only 8.8\%, demonstrating greater robustness across different datasets.

\subsection{Attention Visualization of the FCG}
\label{sec:attn_vis}
\begin{figure}[th]
\centering
\includegraphics[trim={0.9cm 0.7cm 1cm 1cm},clip,width=0.9\linewidth]{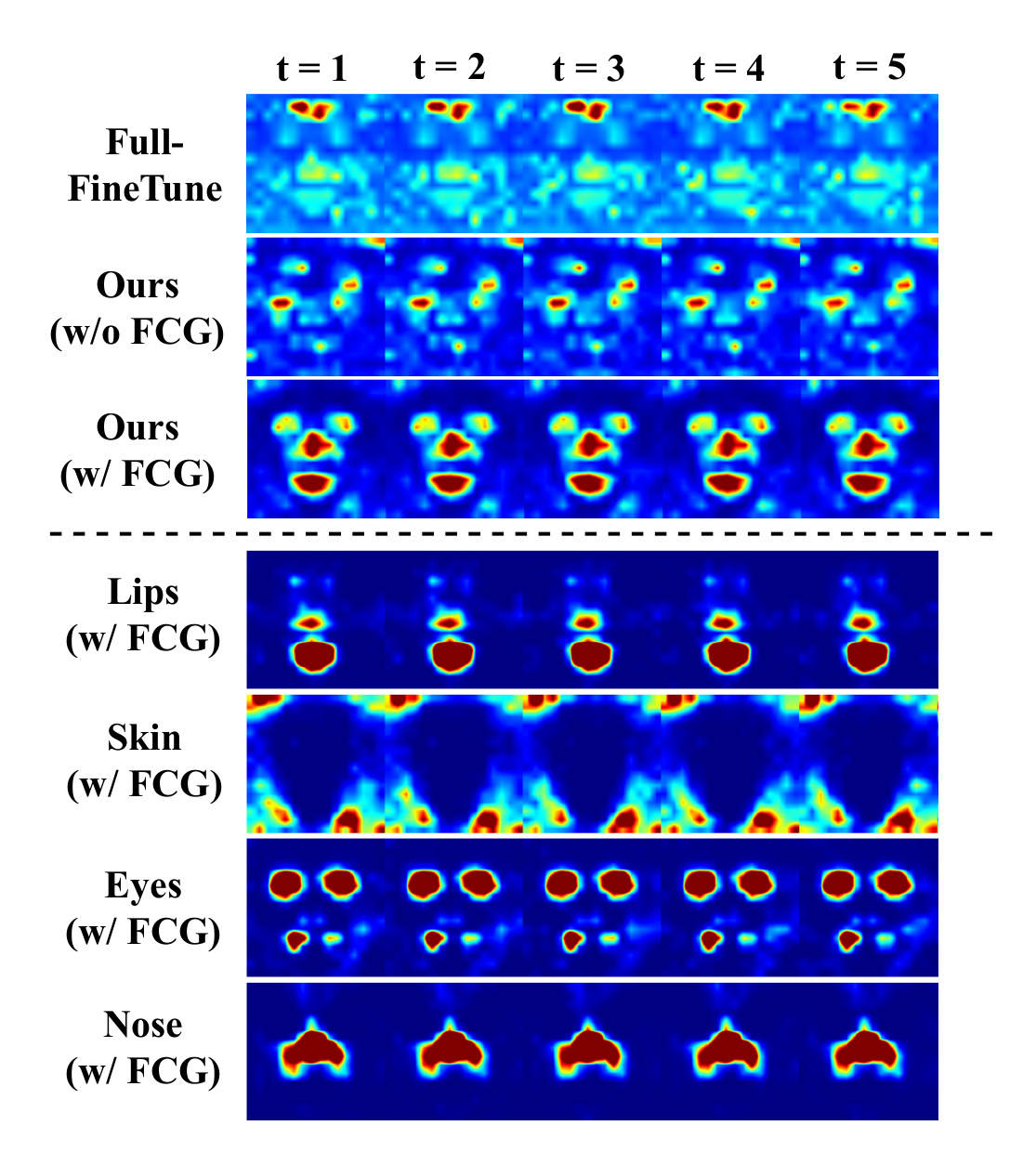} 
\caption{\textbf{Attention Visualization:} The upper section displays the per-frame affinity maps for three Deepfake detectors: the fully fine-tuned CLIP image encoder and our proposed framework with and without the FCG. The lower section further illustrates the per-frame affinity maps for each of the four queries in our framework with the FCG applied. The results demonstrate the effectiveness of our proposed FCG mechanism in steering the model's attention toward different facial components.}
\label{fig:visualize}
\vspace{-10pt}
\end{figure}

\begin{figure*}[t]
\centering
\includegraphics[trim={0.5cm 0.5cm 0.5cm 0.5cm},clip,width=1.0\textwidth]{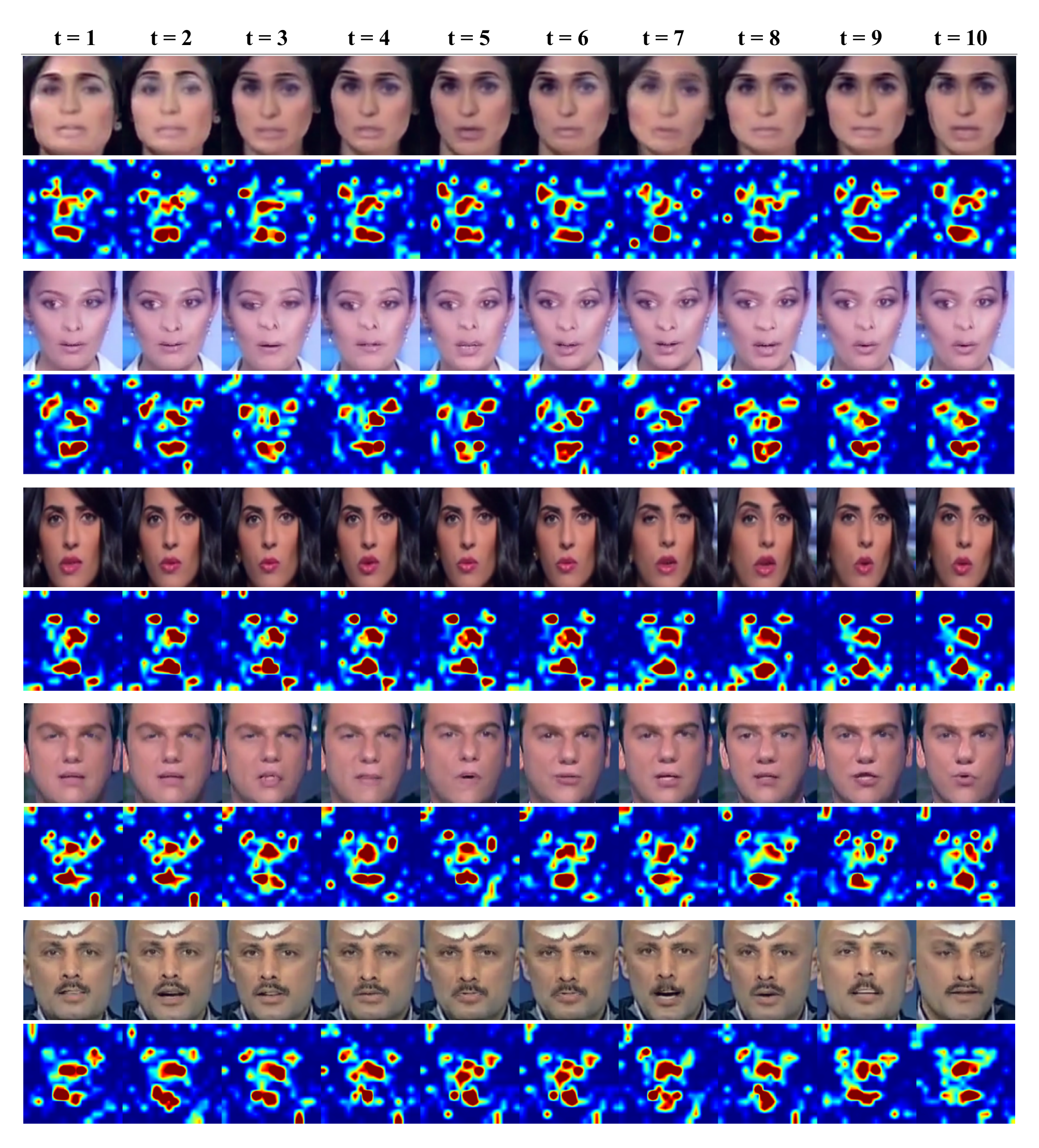} 
\caption{\textbf{Attention Visualization for Individuals:} We present the input frames along with the per-frame attention affinity map for individual subjects. We retain the experimental settings described in Sec.~\ref{sec:attn_vis} while sampling only a single clip for visualization.}
\label{fig:attn_vis_single}
\end{figure*}

In Figure~\ref{fig:visualize}, we present a comparative visualization of the per-frame affinity maps for three models: the fully fine-tuned CLIP image encoder (as shown in Table \ref{table:cross_arch}) and our framework with and without the FCG. {The results presented are averaged over 100 randomly sampled clips from the FF++ testing dataset (please refer to Fig.~\ref{fig:attn_vis_single} for visualizations of individuals
).} The upper part of the figure shows the averaged result when multiple queries are present in the model, while the lower part specifically displays each of the four queries in our framework with the FCG applied. This juxtaposition highlights the effectiveness of our strategy in directing the model's attention toward four key facial regions, whereas the other models tend to focus on arbitrary small areas that lack generalizability for detecting Deepfakes.

\section{Conclusion}
In this paper, we present a novel PEFT framework that leverages the CLIP image encoder to detect Deepfake videos by identifying temporal inconsistencies and spatial artifacts in critical facial regions. We introduce the Facial Component Guidance (FCG) loss, which directs the spatial branch to focus on four key facial areas: skin, nose, eyes, and lips, for enhanced feature extraction. In cross-dataset evaluations, our method outperforms previous state-of-the-art methods across multiple datasets. Our experiments also reveal that prior evaluation methods may overlook the model's bias toward the training dataset. Through extensive experiments, our method demonstrates advantages in parameter efficiency, data efficiency, and robustness.

\section{Limitations and Future Work}
In our framework, we sparsely sample a limited number of frames from each video clip to reduce memory consumption. However, this approach may overlook high-frequency artifacts present in Deepfake videos, potentially leading to incorrect predictions. Despite achieving commendable robustness compared to other methods, we believe that performance could be further improved by enhancing robustness during the pretraining stage. Our framework is designed with high extensibility, allowing for the deployment of different modules in the decoder block to address various aspects of Deepfake detection. Incorporating self-supervised learning methods as additional modules in the decoder block could also be a promising avenue for further development and improved performance.

\section{Acknowledgement}
This research is supported by National Science and Technology Council, Taiwan (R.O.C) under the grant numbers NSTC-113-2634-F-002-007, NSTC-112-2222-E-001-001-MY2, NSTC-113-2634-F-002-008, NSTC-113-2634-F-001-002-MBK, NSTC-111-2221-E-011-128-MY3 and Academia Sinica under the grant number of AS-CDA-110-M09. We thank to National Center for High-performance Computing (NCHC) of National Applied Research Laboratories (NARLabs) in Taiwan for providing computational and storage resources.

{
    \small
    \bibliographystyle{ieeenat_fullname}
    \bibliography{main}
 
}

\clearpage

\maketitlesupplementary

\section{More Experiments for Model Analysis}
In this section, we provide additional experiments to further analyze our framework. We remain the Area Under the receiver operating characteristic Curve (AUC) as the evaluation metric in the experiments.

\subsection{Importance of Facial Components in the FCG}
In Table \ref{table:ablate_facial_part}, we evaluate the cross-dataset performance of the models by excluding specific facial components from our Face Component Guidance (FCG) mechanism. We observe that excluding guidance for any component results in a slight decrease in average performance. Notably, the greatest performance drop occurs when the `eyes' facial component is excluded, suggesting it more critical than other facial components for generalization. Consequently, we include all facial components in our FCG, as this approach achieves the best overall performance across all datasets.

\subsection{The Selection of the Target Attribute}
In Table \ref{table:ablate_ddot_gamma}, we explore different target attribute \({\gamma^{s}}\) settings to evaluate the effectiveness of attributes to improve model generalizability to focus on major facial parts. We can see in the table that selecting \(k\)  and \(q\) both performs well in improving generalizability while \(v\) have a degraded performance. We surmise this due to the design nature of \(q\) and \(k\) in for affinity evaluations. In our experiments, we set \( {\gamma^{s}} = k \) due to its slightly advanced performance.

\subsection{Evaluation on Videos of AI Avatars}
To further validate our model's generalizability to modern AI video generators, we collect a dataset comprising 56 videos from 28 different content creators who use \textbf{HeyGen}\footnote{https://www.heygen.com/} to generate videos with their AI avatars. We evaluate these collected videos using our method, RealForensics~\cite{realforensic}, and LAA-Net (w/ SBI)~\cite{LAA-Net} for comparison. All models are pre-trained on the FF++ dataset, and the scores from previous works are evaluated using their official code and model checkpoints under the default settings. Our method achieves an AUC of 86.1\%, while the RealForensics method achieves an AUC of 84.9\%, and the LAA-Net (w/ SBI) method achieves only 45.4\% AUC. The superior performance of our approach compared to previous SOTA methods further demonstrates its enhanced generalizability.

\begin{table}[t]
\caption{\textbf{Importance of Facial Components in FCG:} We evaluate the cross-dataset performance on models trained with excluded facial components in the FCG. This experiment demonstrates the impact of each facial parts to improve model generalization. }
\vspace{-15pt}
\setlength{\tabcolsep}{5pt} 
\renewcommand{\arraystretch}{1} 
\begin{center}
\label{table:ablate_facial_part}
\begin{tabular}{l| c c c c c| l }
\hline
Method &  CDF & DFDC & FSh & DFo & WDF & Avg. \\
\hline
\textbf{Ours} & \textbf{95.0} & \textbf{81.8} & \textbf{98.1} & \textbf{99.6}  & \textbf{87.2} & \textbf{92.3}  \\
\hline
w/o eyes & 93.9& 81.2& 98.1 & 99.3 & 85.8 & 91.5 \\ 
w/o nose & 94.3 & \underline{81.7} & 97.1 & 99.3 & 86.0 & 91.7 \\ 
w/o lips & \underline{94.5} & 81.5 & \underline{97.8} & \textbf{99.6} & 86.7 & \underline{92.0} \\ 
w/o skin & 94.1 & 81.6 & 97.4 & \underline{99.5} & \underline{87.1} &91.9 \\ 
\hline
\end{tabular}
\end{center}
\vspace{-10pt}
\end{table}

\begin{table}[t]
\setlength{\tabcolsep}{5pt} 
\renewcommand{\arraystretch}{1.2} 
\caption{\textbf{Evaluation of \({\gamma^{s}}\) Parameter}: We select different \({\gamma^{s}}\) values in \(\{q, k, v\}\) to evaluate the efficacy for attributes to collect informative facial features.}
\vspace{-15pt}
\begin{center}
\label{table:ablate_ddot_gamma}
\begin{tabular}{  c | c c c c c| c  }
\hline
\({\gamma^{s}}\)    &   CDF & DFDC &FSh & DFo & WDF & Avg.\\
\hline
\({\gamma^{s}} = k\)	& \underline{95.0} & \textbf{81.8} & \textbf{98.1} & \textbf{99.6} & \textbf{87.2}& \textbf{92.3} \\
\hline
\({\gamma^{s}} = q\) &	\textbf{95.1} & 81.5 &  97.8 & 99.2  & 86.7 &\underline{92.1} \\
\({\gamma^{s}} = v\) &	\underline{84.2} & \underline{81.6} &  \underline{97.9} & \underline{99.5} & 86.0 & 89.8 \\
\hline 
\end{tabular}
\end{center}
\vspace{-15pt}
\end{table}

\section{Evaluation on Modern Deepfake Techniques}
Beyond comprehensive comparisons on video-based Deepfake detection, we also evaluate the proposed approach on unseen novel Deepfakes, with a particular focus on recent Diffusion models. Since the latest advancements in Deepfakes using Diffusion models are image-based, we adapt our video-based framework to operate under similar conditions by removing the temporal module and retaining only the spatial module in the decoder block. 

 To ensure fair comparison, we followed the protocol from DIRE~\cite{dire} and utilize the CelebA-HQ subset from their proposed DiffusionForensics~\cite{dire} dataset, which contains facial images generated by Diffusion Models (DMs). It includes real images from CelebA-HQ~\cite{celebahq} and fake images generated by SD-v2~\cite{sdv2} as the training subset, while the testing subset further includes images generated from IF~\cite{IF}, DALLE-2~\cite{DALLE2}, and Midjourney.
The results of the following experiments are reported using the Average Precision (AP) metric, expressed as a percentage.

\vspace{5pt}
\noindent\textbf{Generalizability to Diffusion Models.} In Table~\ref{table:gen_face_evaluation}, we evaluate the effectiveness of our framework on images generated by novel diffusion models (DMs). In the upper section, we assess the zero-shot capability of our framework alongside the State-Of-The-Art (SOTA) image-based SBI method. Both methods are pre-trained on the FF++ dataset and evaluated on the testing subset from DiffusionForensics. The results demonstrate that our framework exhibits stronger zero-shot capability, outperforming the SOTA SBI by an average of 15.8\% AP. This performance can be attributed to the FCG, which prevents the model from overfitting to dataset-specific cues.

In the lower section, we follow the protocols of previous methods to train and evaluate our framework on the DiffusionForensics dataset. We compare our approach against prior methods (CNNDet~\cite{wang2020}, F3Net~\cite{f3net}, and DIRE~\cite{dire}) to demonstrate its generalizability. The results in the table show that our framework achieves performance on par with the SOTA method (DIRE).

\begin{table}[t]
\begin{center}
\setlength{\tabcolsep}{3pt} 
\renewcommand{\arraystretch}{1} 
\caption{\textbf{Evaluation on Novel Diffusion Deepfakes}: In the upper table, we evaluate the zero-shot capability of our framework with the SBI. In the bottom table, we compare with methods trained on the CelebA-HQ split of the DiffusionForensics dataset.}
\label{table:gen_face_evaluation}
   {%
    \begin{tabular}{lccccc}
    \toprule \multirow{2}{*}{ Method } & \multicolumn{5}{c}{ Generated face images } \\
\cline { 2 - 6 } & SD-v2 & IF & DALLE-2 & Midjourney & Avg. \\
\hline 
SBI~\cite{sbi} & 70.8 & 83.9 & 64.4  & 41.5 & 65.2\\
\textbf{Ours} & \textbf{96.8} & \textbf{93.1} & \textbf{71.4}  & \textbf{62.7} & \textbf{81.0} \\   \hline 
CNNDet~\cite{wang2020}  & \underline{99.8} & 82.7 & 33.7  & 69.3  & 71.4  \\
F3Net~\cite{f3net}  & 99.1 & 84.9 & 69.8  & \underline{87.9}  & \underline{85.4}  \\
DIRE~\cite{dire} & \textbf{100}  & \underline{99.9} & \textbf{99.9} & \textbf{100}  & \textbf{100}   \\
 \textbf{Ours} & \textbf{100} & \textbf{100} & \underline{99.8}  & \textbf{100} & \textbf{100} 
 \\
\bottomrule
\end{tabular}%
 }
\end{center}
\vspace{-15pt}
\end{table}

\vspace{5pt}
\noindent\textbf{Robustness Towards Common Perturbations.} In real-world scenarios, images often undergo various post-processing adjustments, making robustness to unseen perturbations crucial. In this section, we evaluate the robustness of our framework against two types of disturbances: Gaussian blur (\(\sigma = 0, 1, 2, 3\)) and JPEG compression (quality = 100, 65, 30). We follow the evaluation setup from the previous section to assess robustness under both zero-shot and in-domain regimes. The results are presented in Table~\ref{table:gen_face_robustness_evaluation}. Our model demonstrates strong robustness, with no significant performance degradation under these perturbations, particularly in the zero-shot evaluation.

\section{Elaboration on Attribute Extraction}
\label{sec:elab_attn_attr_extract}
To elucidate the connection between attention attributes and patch embeddings within the Vision Transformer (ViT) encoder pipeline, we detail the workflow of a typical ViT encoder layer in Algorithm \ref{alg:vit_layer}. Each layer accepts embeddings from the preceding layer as input, which encompasses a class embedding along with numerous patch embeddings. These embeddings are then processed through a self-attention mechanism to produce output embeddings for the subsequent layer. Initially, the class embedding is represented by a learnable token, and patch embeddings are formed by a distinct patch extraction layer given an image (for further details, please see the ViT paper~\cite{ViT}). In the algorithm presented, \(W_{s}\) and \(B_{s}\), for \(s \in \{Q,K,V,O\}\), signify the weights and biases associated with the linear transformations. \(LN_{1}\) and \(LN_{2}\) represent the layer normalization modules. The \textbt{MHSA} stands for the Multi-Head Self-Attention mechanism, which operates on the query, key, and value embeddings of the class and patch tokens. Furthermore, the \(MLP\) (multi-layer perceptron) includes two linear layers and a GeLU activation layer. The attention attributes \(\mathcal{A}_{l,\gamma}\), where \( \gamma \in \{ q, k, v \}\), are extracted as specified in the cited lines \ref{alg:attr_q}, \ref{alg:attr_k}, and \ref{alg:attr_v}, and the extracted patch embeddings \(\mathcal{P}_l\) are referred to in line \ref{alg:attr_emb}.

\section{Inference Time}
The average inference time of our framework on a 3-second video is 1.5 seconds using an A5000 GPU. As our framework leverages the CLIP image encoder to extract generic features for adaptation, most of the inference time is spent in the image encoder processing the 10 frames extracted from the video clip. In contrast, our proposed lightweight decoder modules require minimal inference time.

\begin{table}[t]
\begin{center}
\setlength{\tabcolsep}{2pt} 
\renewcommand{\arraystretch}{1} 
\caption{\textbf{Robustness on Novel Diffusion Deepfakes:} We assess the zero-shot and in-domain robustness of our framework with SBI and DIRE, respectively.}
\label{table:gen_face_robustness_evaluation}
{%
\begin{tabular}{lcccccccc}
\toprule 
\multirow{2}{*}{ Method } & \multicolumn{3}{c}{ JPEG (Quality) } & \multicolumn{4}{c}{ Blur (Sigma) }  & \multirow{2}{*}{ Avg. } \\
\cmidrule(lr){2-4} \cmidrule(lr){5-8} & 100 & 65 & 30 & 0 & 1 & 2 & 3 \\
\hline 
SBI & 65.2 & 53.5 & 56.3  & 65.2 & 51.3 & 52.9 & 54.2 & 55.6\\
\textbf{Ours} & \textbf{100} & \textbf{79.3} & \textbf{76.2}  & \textbf{100} & \textbf{81.5} & \textbf{81.1} & \textbf{79.2} & \textbf{85.3} \\   
\hline 
DIRE* & \textbf{100}  & 99.8 & 99.8 & \textbf{100}  & \textbf{99.9} & \textbf{99.9} & \textbf{99.9} & \textbf{99.9}
\\
\textbf{Ours}* & \textbf{100} & \textbf{100} & \textbf{99.9}  & \textbf{100} & \textbf{99.9} & \textbf{99.9} & \textbf{99.9} & \textbf{99.9}
\\
\bottomrule
\end{tabular}%
}
\end{center}
\vspace{-10pt}
\end{table}

\begin{figure}[t]
\centering
\begin{minipage}{1.0\linewidth}
\vspace{-10pt}
\begin{algorithm}[H]
    \caption{The workflow of the ViT transformer encoder layer given embeddings from the previous layer.}\label{alg:vit_layer}
    \begin{algorithmic}[1]
        \State \textbf{Input} $x$
        \State \textbf{Output} $emb$
        \State $\hat{x} \gets LN_{1}(x)$
        \State $q \gets W_Q\hat{x}+B_Q\hat{x}$ \label{alg:attr_q} \Comment{Query Transform}
        \State $k \gets W_K\hat{x}+B_K\hat{x}$ \label{alg:attr_k} \Comment{Key Transform}
        \State $v \gets W_V\hat{x}+B_V\hat{x}$ \label{alg:attr_v} \Comment{Value Transform}
        \State $z \gets \textbt{MHSA}(q,k,v)$  \Comment{Multi-head Self-Attention}
        \State $out \gets W_Oz + B_Oz$ \label{alg:attr_out}
        \State $x' \gets x + out$
        \State $emb \gets x' + MLP(LN_{2}(x'))$ \label{alg:attr_emb}
    \end{algorithmic}
\end{algorithm}
\end{minipage}
\end{figure}

\section{Societal Impact Concern}
Since the proposed method mainly works on Deepfake detection problem to mitigate the negative influences brought by Deepfake technologies, there is no major societal impact concerns.

\end{document}